\documentclass[runningheads]{llncs}
\usepackage{graphicx}

\usepackage{tikz}
\usepackage{comment}
\usepackage{amsmath,amssymb} 
\usepackage{color}
\usepackage[accsupp]{axessibility}  

\usepackage{booktabs, siunitx, etoolbox}
\usepackage{hyperref}
\usepackage{bm}
\usepackage{mathtools}
\usepackage{cite}
\usepackage{multirow}
\usepackage{subcaption}
\usepackage{floatrow}

\usepackage[capitalize]{cleveref}
\crefname{section}{Sec.}{Secs.}
\Crefname{section}{Section}{Sections}
\Crefname{table}{Table}{Tables}
\crefname{table}{Tab.}{Tabs.}

\newfloatcommand{capbtabbox}{table}[][\FBwidth]
\DeclareMathOperator*{\argmin}{argmin}
\DeclareMathOperator*{\argmax}{argmax}

\usepackage[width=122mm,left=12mm,paperwidth=146mm,height=193mm,top=12mm,paperheight=217mm]{geometry}

\begin{document}
\pagestyle{headings}
\mainmatter
\title{Zero-Shot Category-Level Object Pose Estimation}
\titlerunning{Zero-Shot Category-Level Object Pose Estimation}
\author{Walter Goodwin
 \inst{1}\thanks{These authors contributed equally} \and
Sagar Vaze \inst{2}$^\star$ \and
Ioannis Havoutis \inst{1} \and
Ingmar Posner \inst{1}}
\authorrunning{W. Goodwin et al.}
\institute{Oxford Robotics Institute, University of Oxford \and
Visual Geometry Group, University of Oxford \\
\email{firstname@robots.ox.ac.uk}\\}
\maketitle
%
\vspace{-8mm}
\begin{abstract}
Object pose estimation is an important component of most vision pipelines for embodied agents, as well as in 3D vision more generally.
In this paper we tackle the problem of estimating the pose of novel object categories in a zero-shot manner. 
This extends much of the existing literature by removing the need for pose-labelled datasets or category-specific CAD models for training or inference.
Specifically, we make the following contributions.
First, we formalise the zero-shot, category-level pose estimation problem and frame it in a way that is most applicable to real-world embodied agents. 
Secondly, we propose a novel method based on semantic correspondences from a self-supervised vision transformer to solve the pose estimation problem.
We further re-purpose the recent CO3D dataset to present a controlled and realistic test setting.
Finally, we demonstrate that all baselines for our proposed task perform poorly, and show that our method provides a \textit{six-fold improvement} in average rotation accuracy at 30 degrees. Our code is available at \url{https://github.com/applied-ai-lab/zero-shot-pose}.
\end{abstract}

\vspace{-3 em}
\begin{figure}[tbh]
\centering
  \includegraphics[width=0.98\columnwidth]{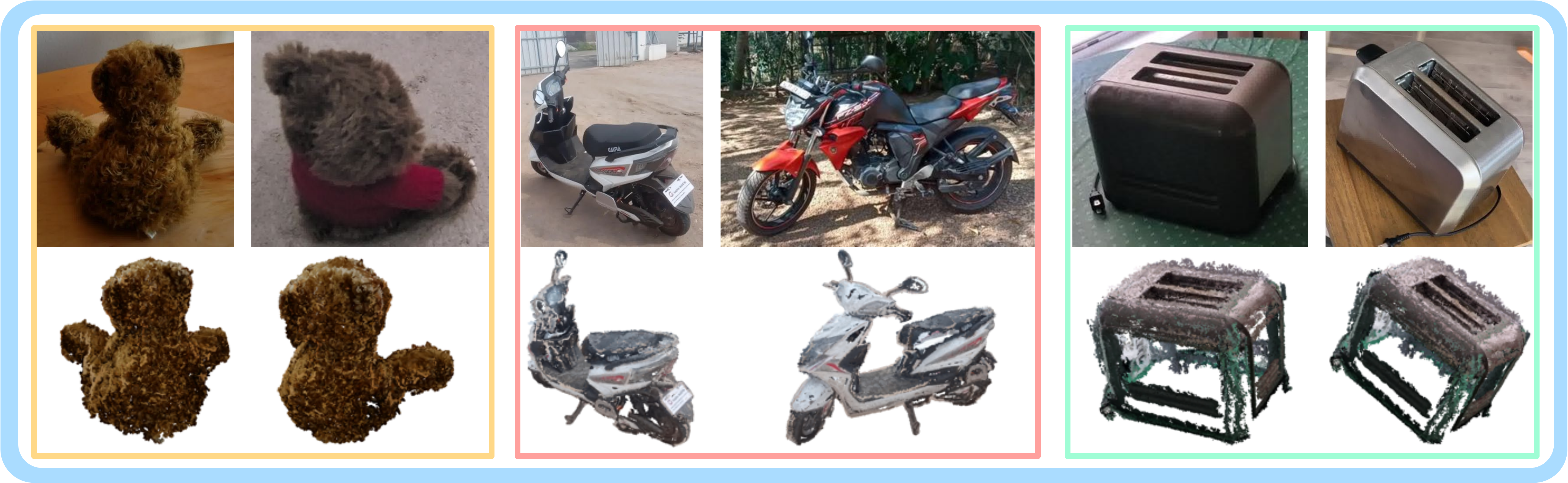}
  \caption{Zero-shot category-level pose estimation enables the alignment of different instances of the same object category, without any pose labels for that category or any other. For each category, the estimated pose of the first object, relative to the second, is visualised through transformation of the first object's point cloud.}
  \vspace{-1em}
  \label{fig:teaser}
\end{figure}
\vspace{-2.6em}
\section{Introduction} \label{sec:intro}
Consider a young child who is presented with two toys of an object category they have never seen before: perhaps, two toy aeroplanes. 
Despite having never seen examples of `aeroplanes' before, the child has the ability to understand the spatial relationship between these related objects, and would be able to align them if required. 
This is the problem we tackle in this paper: the zero-shot prediction of pose offset between two instances from an object category, without the need for any pose annotations.
We propose this as a challenging task which removes many assumptions in the current pose literature, and which more closely resembles the setting encountered by embodied agents in the real-world.
To substantiate this claim, consider the information existing pose recognition algorithms have access to. 
Current methods make one (or more) of the following assumptions:
that evaluation is performed at the \textit{instance-level} (i.e there is no intra-category variation between objects) \cite{Xiang2018}; that we have access to \textit{labelled pose datasets} for all object categories \cite{GuptaPose2015, Xiao2021, ChenFSNet2021, Akizuki2021, Pavlakos2017, Zhou2018, Wang2019NOCS}; and/or that we have access to a realistic \textit{CAD model} for each object category the model will encounter \cite{Grabner2018, ChenSGPA2021, Xiao2019}.

Meanwhile, humans can understand pose without access to any of this information. 
How is this possible?
Intuitively, we suggest humans use an understanding of \textit{object parts}, which generalise across categories, to correspond related objects.
This process can be followed by using \textit{basic geometric primitives} to understand the spatial relationship between objects.
Humans typically also have a coarse \textit{depth estimate} and can inspect the object from \textit{multiple viewpoints}.

In this paper, we use these intuitions to build a solution to estimate the pose offset between two instances of a given category. 
We perform `zero-shot' pose-estimation in the sense that our models have never seen pose-labelled examples of the test categories, and neither do they rely on category-specific CAD models.
We first make use of features extracted from a vision transformer (ViT \cite{dosovitskiy2020vit}), trained in a self-supervised manner on large scale data \cite{Caron2021}, to establish semantic correspondences between two object instances of the same category.
Prior work has demonstrated that self-supervised ViTs have an understanding of object parts which can transfer to novel instances and categories \cite{vaze22gcd, Amir2021}.
Next, using a weighting of the semantic correspondences, we obtain a coarse estimate of the pose offset by selecting an optimal viewpoint for one of the object instances. 
Having obtained semantic correspondences and selected the best view, we use depth maps to create sparse point clouds for each object at the corresponding semantic locations. 
Finally, we align these point clouds with a rigid-body transform using a robust least squares estimation \cite{Umeyama1991} to give our final pose estimate.

We evaluate our method on the CO3D dataset \cite{ReizensteinCO3D2021}, which provides high-resolution imagery of diverse object categories, with substantial intra-category appearance differences between instances.
We find that this allows us to reflect a realistic setting while performing quantitative evaluation in a controlled manner.
We consider and compare to a range of baselines which could be applied to this task, but find that they
perform poorly and often fail completely, demonstrating the highly challenging nature of the problem. 

In summary, we make the following contributions:
\begin{itemize}
    \item We formalise a new and challenging setting for pose estimation, which is an important component of most 3D vision systems. We suggest our setting closely resembles those encountered by real-world embodied agents (\cref{sec:formalize_zero_shot_pose}).
    \item We propose a novel method for zero-shot, category-level pose estimation, based on semantic correspondences from self-supervised ViTs (\cref{sec:methods}).
    \item Through rigorous experimentation on a devised CO3D benchmark, we demonstrate that our method facilitates zero-shot pose alignment when the baselines often fail entirely (\cref{sec:exps}).
\end{itemize}
\section{Related Work} \label{sec:related_work}
\subsection{Category-level pose estimation}
While estimating pose for a single object instances has a long history in robotics and computer vision \cite{Xiang2018}, in recent years there has been an increased interest in the problem of \textit{category-level} pose estimation, alongside the introduction of several category-level datasets with labelled pose \cite{Xiang2014, Xiang2016, AhmadyanObjectron2021, Wang2019NOCS}. Approaches to category-level pose estimation can be broadly delineated into: those defining pose explicitly through the use of reference CAD models \cite{Shi2021, Xiao2019, Sahin2019, Shi2021, Grabner2018}; those which learn category-level representations against which test-time observations can be in some way matched to give relative pose estimates \cite{Pavlakos2017, Zhou2018, Wang2019NOCS, Chen2020, ChenSGPA2021, ChenCASS2020, Wang2021_NeMo, Tian2020}; and those that learn to directly predict pose estimates for a category from observations \cite{GuptaPose2015, Xiao2021, ChenFSNet2021, Akizuki2021}. 

Most methods (e.g. \cite{Tian2020, Wang2019NOCS, Pavlakos2017, ChenSGPA2021, Lin2021Pose}) treat each object category distinctly, either by training a separate model per category, or by using different templates (e.g. CAD models) for each category. A few works (e.g. \cite{Zhou2018, Xiao2021}) attempt to develop category-agnostic models or representations, and several works consider the exploitation of multiple views to enhance pose estimation \cite{Kanezaki2018, Kundu2019}. In contrast to existing works in category-level pose estimation, we do not require any pose-labelled data or CAD models in order to estimate pose for a category, and tackle pose estimation for unseen categories.

\subsection{Few-shot and self-supervised pose estimation}
There has been some recent work that notes the difficulty of collecting large, labelled, in-the-wild pose datasets, and thus seeks to reduce the data burden by employing few-shot approaches. For instance, Pose-from-Shape \cite{Xiao2019} exploits existing pose-labelled RGB datasets, along with CAD models, to train an object-agnostic network that can predict the pose of an object in an image, with respect to a provided CAD model. Unlike this work, we seek to tackle an in-the-wild setting in which a CAD model is not available for the objects encountered. Self-supervised, embodied approaches for improving pose estimation for given object instances have been proposed \cite{Deng2020}, but require extensive interaction and still do not generalise to the category level. Few-shot approaches that can quickly fine-tune to previously unseen categories exist \cite{Tseng2019, Xiao2021}, but still require a non-trivial number of labelled examples to fine-tune to unseen categories, while in contrast we explore the setting in which no prior information is available. Furthermore, recent works have explored the potential for unsupervised methods with equivariant inductive biases to infer category-level canonical frames without labels \cite{Li2021SE3, Sajnani2022}, and to thus infer 6D object pose given an observed point cloud. These methods, while avoiding the need for pose labels, only work on categories for which they have been trained.
Finally, closest in spirit to the present work is \cite{Banani2020}, who note that the minimal requirement to make zero-shot pose estimation a well-posed problem is to provide an implicit canonical frame through use of a reference image, and formulate pose estimation as predicting the relative viewpoint from this view. However, this work can only predict pose for single object instances, and does not extend to the category level.

\subsection{Semantic descriptor learning}
A key component of the presented method to zero-shot category level pose estimation is the ability to formulate semantic keypoint correspondences between pairs of images within an object category, in a zero-shot manner. There has been much interest in semantic correspondences in recent years, with several works proposing approaches for producing these without labels \cite{Lee2019SFNET, Aberman2018, Amir2021, Caron2021}. 
Semantic correspondence is particularly well motivated in robotic settings, where problems such as extending a skill from one instance of an object to any other demand the ability to relate features across object instances. Prior work has considered learning dense descriptors from pixels \cite{Florence2018} or meshes \cite{Simeonov2021} in a self-supervised manner, learning skill-specific keypoints from supervised examples \cite{Manuelli2019}, or robust matching at the whole object level \cite{Goodwin2021}. The descriptors in \cite{Florence2018, Simeonov2021, Manuelli2019} are used to infer the relative pose of previously unseen object instances to instances seen in skill demonstrations. In contrast to these robotics approaches, in our method we leverage descriptors that are intended to be category-agnostic, allowing us to formulate a zero-shot solution to the problem of pose estimation.
\section{Zero-shot Category-Level Pose Estimation} \label{sec:formalize_zero_shot_pose}
\begin{figure}[h]
\centering
  \includegraphics[width=0.98\columnwidth]{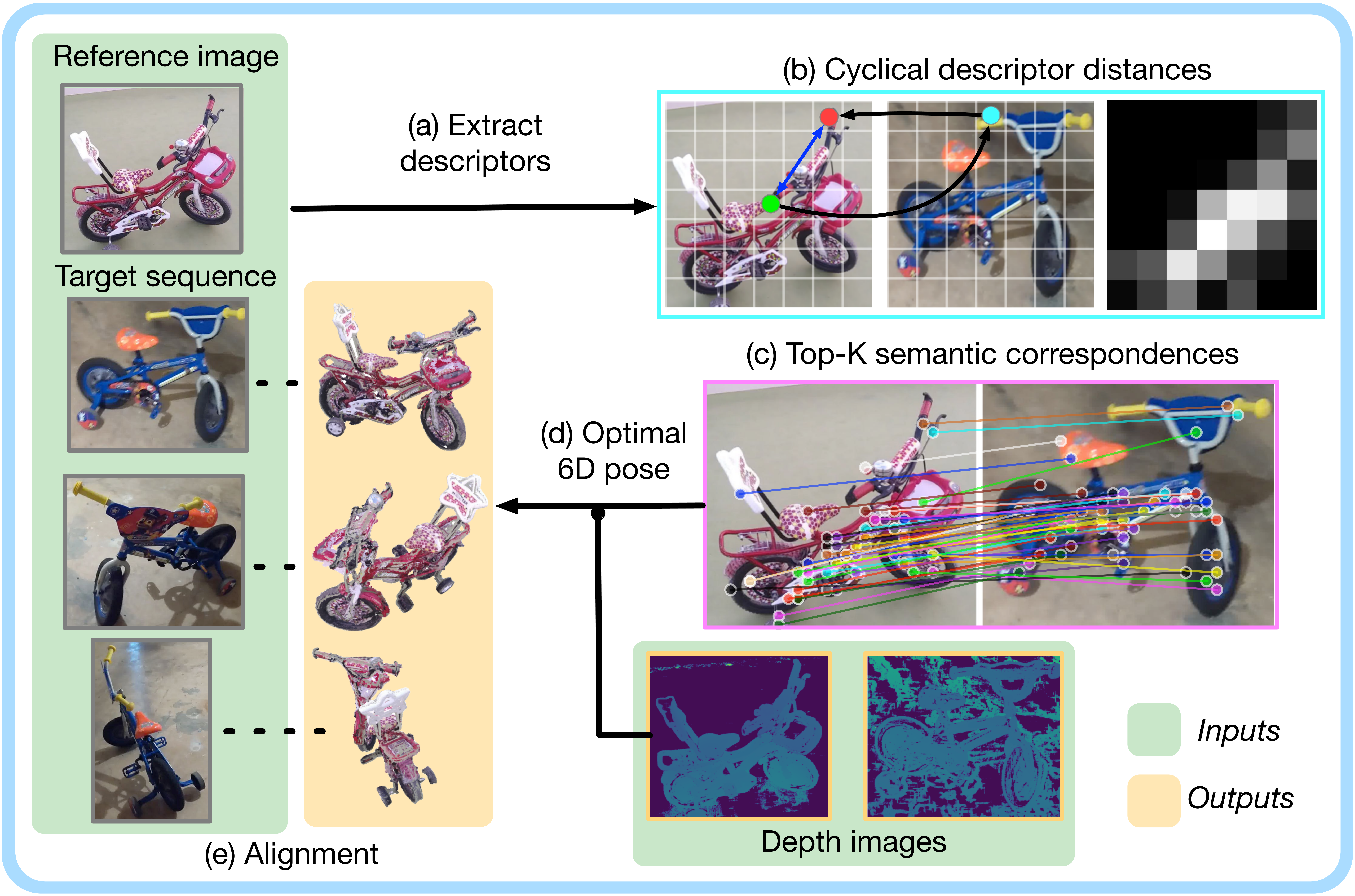}
  \caption{Our method for zero-shot pose estimation between two instances of an object, given a reference image and a sequence of target images.
  In our method, we: \textbf{(a)} Extract spatial feature descriptors for all images with a self-supervised vision transformer (ViT). 
  \textbf{(b)} Compare the reference image to all images in the target sequence by building a set of cyclical distance maps (\cref{sec:semantic_correspond}). 
  \textbf{(c)} Use these maps to establish $K$ semantic correspondences between compared images and select a suitable view from the target sequence (\cref{sec:best_view}). 
  \textbf{(d)} Given the semantic correspondences and a suitable target view, we use depth information to compute a rigid transformation between the reference and target objects (\cref{sec:pose_computation_with_depth}).
  \textbf{(e)} Given relative pose transformations between images in the target sequence, we can align the point cloud of the \textit{reference image} with \textit{the entire target sequence}.}
  \label{fig:main}
\end{figure}
\vspace{-2mm}

In this section, we formalise and motivate our proposed zero-shot pose estimation setting.
To do this, we first outline the generic problem of object pose estimation. 6D object pose estimation entails estimating the offset (translation and rotation) of an object with respect to some frame of reference, normally given just an image of the object. 
This frame of reference can be defined implicitly (e.g in the supervised setting, the labels are all defined with respect to some `canonical' frame) or explicitly (e.g with a reference image). 
In either case, pose estimation is fundamentally a relative problem. 
In the zero-shot setting we consider, the frame of reference cannot be implicitly defined by labels: we do not have labelled pose for any objects.
Therefore, the pose estimation problem is that of aligning (computing the pose offset between) two instances of a given category, i.e. a reference and a target object.

In our proposed setting, we assume access to $N$ views of a target object, as well as depth information for both objects, and suggest that these constraints reflect practical settings.
For objects in the open-world, we are unlikely to have realistic CAD models or labelled pose training sets.
On the other hand, many embodied agents are fitted with depth cameras or can recover depth (up to a scale) from structure from motion or stereo correspondence.
Furthermore, real-world agents are able to interact with the object and hence gather images from multiple views.

Formally, we consider a reference image, $I_{\mathcal{R}}$, and a set of target images $I_{\mathcal{T}_{1:N}} = \{I_{\mathcal{T}_1}...I_{\mathcal{T}_N}\}$, where $I_i \in \mathbb{R}^{H \times W \times 3}$. 
We further have access to depth maps, $D_i \in \mathbb{R}^{H \times W}$ for all images.
Given this information, we require a model, $\mathcal{M}$, to output a single 6D pose offset between the object in the reference image and the object in the target sequence, as:

\begin{equation}
    T^{*} = \mathcal{M}(I_{\mathcal{R}}, I_{\mathcal{T}_{1:N}} | \quad D_\mathcal{R}, D_{\mathcal{T}_{1:N}})
\end{equation}

Finally, we note that, in practice, the transformations between the target views must be known for the predicted pose offset to be most useful.
These transformations are easily computed by an embodied agent and can be used to align the reference instance with the entire target sequence, given an alignment between $I_{\mathcal{R}}$ and \textit{any} of the target views.

\section{Methods} \label{sec:methods}

In this section, we detail our method for zero-shot pose estimation.
First, semantic correspondences are obtained between the reference and target object (\cref{sec:semantic_correspond}). 
These correspondences are used to select a suitable view for pose estimation from the $N$ images in the target sequence (\cref{sec:best_view}).
Finally, using depth information, the correspondences' spatial locations are used to estimate the pose offset between the reference and target object instances (\cref{sec:pose_computation_with_depth}).

\subsection{Self-supervised semantic correspondence with cyclical distances}
\label{sec:semantic_correspond}

The key insight of our method is that semantic, parts-based correspondences generalise well between different object instances within a category, and tend to be spatially distributed in similar ways for each such object. 
Indeed, a parts-based understanding of objects can also generalise between categories; for instance, `eyes', `ears' and `nose' transfer between many animal classes.  
Recent work has demonstrated that parts-based understanding emerges naturally from self-supervised vision transformer (ViT) features \cite{vaze22gcd, Amir2021, Caron2021}, and our solution leverages such a network with large scale pre-training \cite{Caron2021}. 
The ViT is trained over ImageNet-1K, and we assume that it carries information about a sufficiently large set of semantic object parts to generalise to arbitrary object categories. 

As described in \cref{sec:formalize_zero_shot_pose}, the proposed setting for pose estimation considers a relative problem, between a reference object (captured in a single image) and a target object (with potentially multiple views available). We compare two images (for now referred to as $I_1$, $I_2$) by building a `cyclical distance' map for every pixel location in $I_1$ using feature similarities.
Formally, consider $\Phi(I_i) \in \mathbb{R}^{H' \times W' \times D}$ as the normalised spatial feature of an image extracted by a ViT.
Letting $u$ be an index into $\Phi(I_1)$ as $u \in \{1...H'\} \times \{1...W'\}$, we find its cyclical point $u'$ as:

\begin{equation}
    u' = \argmin_w d(\Phi(I_1)_w, \Phi(I_2)_v) \quad | \quad v = \argmin_w d(\Phi(I_1)_u, \Phi(I_2)_w)
\end{equation}

Here $d(\cdot, \cdot)$ is the L2-distance, and a cyclical distance map is constructed as $C \in \mathbb{R}^{H' \times W'}$ with $C_u = -d(u, u')$. 
Using the top-$K$ locations in $C$, we take features from $\Phi(I_1)$ and their nearest neighbours in $\Phi(I_2)$ as correspondences.
This process is illustrated in \cref{fig:main}b. 

The cyclical distance map can be considered as a soft mutual nearest neighbours assignment.
Mutual nearest neighbours \cite{Amir2021} between $I_1$ and $I_2$ return a cyclical distance of zero, while points in $I_1$ with a small cyclical distance can be considered to `almost' have a mutual nearest neighbour in $I_2$. The proposed cyclical distance metric has two key advantages over the hard constraint.
Firstly, while strict mutual nearest neighbours gives rise to an unpredictable number of correspondences,
the soft measure allows us to ensure $K$ semantic correspondences are found for every pair of images. We find having sufficient correspondences is critical for the downstream pose estimation.
Secondly, the soft constraint adds a spatial prior to the correspondence discovery process: features belonging to the same object part are likely to be close together in pixel space.

Finally, following ~\cite{Amir2021}, after identifying an initial set of matches through our cyclical distance method, we use $K$-Means clustering on the selected features in the reference image to recover points which are spatially well distributed on the object. We find that well distributed points result in a more robust final pose estimate (see supplementary).
In practice, we select the top-$2K$ correspondences by cyclical distance, and filter to a set of $K$ correspondences with $K$-Means.

\subsection{Finding a suitable view for alignment}
\label{sec:best_view}

Finding semantic correspondences between two images which view (two instances of) an object from very different orientations is challenging. 
For instance, it is possible that images from the front and back of an object have no semantic parts in common. 
To overcome this, an agent must be able to choose a suitable view from which to establish semantic correspondences. In the considered setting, this entails selecting the best view from the $N$ target images. 
We do this by constructing a correspondence score between the reference image, $I_{\mathcal{R}}$, and each image in the target sequence, $I_{\mathcal{T}_{1:N}}$. 
Specifically, given the reference image and an image from the target sequence, the correspondence score is the sum the of the feature similarities between their $K$ semantic correspondences. 
Mathematically, given a set of $K$ correspondences between the $j^{th}$ target image and the reference, $\{(u^{j}_{k}, v^{j}_{k})\}_{k = 1}^{K}$, this can be written as:
\begin{equation}
    j^{*} = \argmax_{j \in 1:N} \quad  \sum_{k = 1}^{K} - d( \Phi(I_{\mathcal{R}})_{u^{j}_{k}} ,  \Phi(I_{\mathcal{T}_{j}})_{v^{j}_{k}} ) 
\end{equation}

\subsection{Pose estimation from semantic correspondences and depth} \label{sec:pose_computation_with_depth}
The process described in \cref{sec:semantic_correspond} gives rise to a set of corresponding points in 2D pixel coordinates, $\{(u_{k}, v_{k})\}_{k = 1}^{K}$. Using depth information and camera intrinsics, these are unprojected to their corresponding 3D coordinates, $\{(\mathbf{u}_{k}, \mathbf{v}_{k})\}_{k = 1}^{K}$, where $\mathbf{u}_{k}, \mathbf{v}_{k} \in \mathbb{R}^{3}$. In the pose estimation problem, we seek a single 6D pose that describes the orientation and translation of the target object, relative to the frame defined by the reference object. Given a set of corresponding 3D points, there are a number of approaches for solving for this rigid body transform. As we assume our correspondences are both noisy and likely to contain outliers, we use a fast least-squares method based on the singular value decomposition \cite{Umeyama1991}, and use RANSAC to handle outliers. We run RANSAC for up to 1,000 iterations, with further details in supplementary. The least squares solution recovers a 7-dimensional transform: rotation $\mathbf{R}$, translation $\mathbf{t}$, and a uniform scaling parameter $\lambda$, which we found crucial for dealing with cross-instance settings. The least-squares approach minimises the residuals and recovers the predicted 6D pose offset, $T^{*}$ as:

\begin{equation}
    T^{*} = (\mathbf{R}^{*}, \mathbf{t}^{*}) = \argmin_{(\mathbf{R}, \mathbf{t})} \sum_{k = 1}^{K} \mathbf{v}_{k} - (\lambda \mathbf{R}\mathbf{u}_{k}+\mathbf{t})
\end{equation}

\section{Experiments} \label{sec:exps}

\subsection{Evaluation Setup} \label{sec:datasets}

\paragraph{Dataset, CO3D \cite{ReizensteinCO3D2021}:} To evaluate zero-shot, category-level pose estimation methods, a dataset is required that provides images of multiple object categories, with a large amount of intra-category instance variation, and with varied object viewpoints. 
The recently released Common Objects in 3D (CO3D) dataset fulfils these requirements with 1.5 million frames, capturing objects from 50 categories, across nearly 19k scenes \cite{ReizensteinCO3D2021}. For each object instance, CO3D provides approximately 100 frames taken from a 360º viewpoint sweep with handheld cameras, with labelled camera pose offsets.
The proposed method makes use of depth information, and CO3D provides estimated object point clouds, and approximate depth maps for each image, that are found by a Structure-from-Motion (SfM) approach applied over the sequences \cite{Schoenberger2016}.
We note that, while other object pose datasets exist \cite{Xiang2014, Xiang2018, AhmadyanObjectron2021}, we find them to either be lacking in necessary meta-data (e.g no depth information), have little intra-category variation (e.g be instance level), contain few categories, or only provide a single image per object instance. 
We expand on dataset choice in the supplementary.

\paragraph{Labels for evaluation:} While the proposed pose estimation method requires no pose-labelled images for training, we label a subset of sequences across the CO3D categories for quantitative evaluation. 
We do this by assigning a category-level canonical frame to each selected CO3D sequence.
We exclude categories that have infinite rotational symmetry about an axis (e.g `apple') or have an insufficient number of instances with high quality point clouds (e.g `microwave').
For the remaining 20 categories, we select the top-10 sequences based on a point cloud quality metric.
Point clouds are manually aligned within each category with a rigid body transform.
As CO3D provides camera extrinsics for every frame in a sequence with respect to its point cloud, these alignments can be propagated to give labelled category-canonical pose for every frame in the chosen sequences. Further details are in the supplementary.

\paragraph{Evaluation setting:} For each object category, we sample 100 combinations of sequence pairs, between which we will compute pose offsets. 
For the first sequence in each pair, we sample a single reference frame, $I_\mathcal{R}$, and from the second we sample $N$ target frames, $I_{\mathcal{T}_{1:N}}$. 
We take $N = 5$ as our standard setting, with results for different numbers of views in \cref{tab:ablation_n_tgt} and the supplementary.
For each pair of sequences, we compute errors in pose estimates between the ground truth and the predictions. For the rotation component, following standard practise in the pose estimation literature, we report the median error across samples, as well as the accuracy at 15º and 30º, which are given by the percentage of predictions with an error less than these thresholds. Rotation error is given by the geodesic distance between the ground truth and predicted rotation \cite{Huynh09}.

\paragraph{`Zero-shot' pose estimation:} In this work, we leverage models with large-scale, self-supervised pre-training.
The proposed pose estimation method is `zero-shot' in the sense that it does not use labelled examples (either pose labels or category labels) for any of the object categories it is tested on. The self-supervised features, though, may have been trained on images containing unlabelled instances of some object categories considered.
To summarise, methods in this paper \textbf{do not require} labelled pose training sets or CAD models for the categories they encounter during evaluation.
They \textbf{do require} large-scale unsupervised pre-training, depth estimates, and multiple views of the target object. 
We assert that these are more realistic assumptions for embodied agents (see \cref{sec:formalize_zero_shot_pose}).

\subsection{Baselines} \label{sec:baselines}

We find very few baselines in the literature which can be applied to the highly challenging problem of pose-detection on unseen categories.
Though some methods have tackled the zero-shot problem before, they are difficult to translate to our setting as they require additional information such as CAD models for the test objects. We introduce the baselines considered.

\paragraph{PoseContrast \cite{Xiao2021}}: This work seeks to estimate 3D pose (orientation only) for previously unseen categories. The method trains on pose-labelled images and assumes unseen categories will have both sufficiently similar appearance and geometry, and similar category-canonical frames, to those seen in training.
We adapt this method for our setting and train it on 86 of the 100 categories from the ObjectNet3D dataset \cite{Xiao2020} (removing 14 categories that are present in our CO3D setting, to ensure a zero-shot comparison). 
During testing, we extract global feature vectors for the reference and target images with the model, and use feature similarities to select a suitable view.
We then run the PoseContrast model on the reference and selected target image, with the model regressing to an Euler angle representation of 3D pose. PoseContrast estimates pose for each image independently, implicitly inferring the canonical frame for the test object category. We thus compute the difference between the pose predictions for the reference and chosen target image to arrive at a relative pose estimate.

\paragraph{Iterative Closest Point (ICP):} ICP is a point cloud alignment algorithm that assumes no correspondences are known between two point clouds, and seeks an optimal registration. We use ICP to find a 7D rigid body transform (scale, translation and rotation, as in \cref{sec:pose_computation_with_depth}) between the reference and target objects.
We use the depth estimates for each image to recover point clouds for the two instances, aggregating the $N$ views in the target sequence for a maximally complete target point cloud. We use these point clouds with ICP.
As ICP is known to perform better with good initialisation, we also experiment with initialising it from the coarse pose estimate given by our `best view'  method (see \cref{sec:best_view}) which we refer to as `ICP + BV'.

\paragraph{Image Matching:} Finally, we experiment with other image matching techniques.
In the literature, cross-instance correspondence is often tackled by learning category-level keypoints.
However, this usually involves learning a different model for each category, which defeats the purpose of our task.
Instead, we use category-agnostic features and obtain matches with mutual nearest neighbours between images, before combining the matches' spatial locations with depth information to compute pose offsets (similarly to \cref{sec:pose_computation_with_depth}).
We experiment both with standard SIFT features \cite{Lowe2004SIFT} and deep features extracted with an ImageNet self-supervised ResNet-50 (we use SWaV features \cite{caron2020unsupervised}).
In both cases, we select the best view using the strength of the discovered matches between the reference and target images (similarly to \cref{sec:best_view}).

\subsection{Implementation Details}

In this work we use pre-trained DINO ViT features \cite{Caron2021} to provide semantic correspondences between object instances.
Specifically, we use ViT-Small with a patch size of 8, giving feature maps at a resolution of $28 \times 28$ from square $224 \times 224$ images.
Prior work has shown that DINO ViT features encode information on generalisable object parts and correspondences \cite{vaze22gcd, Amir2021}.
We follow \cite{Amir2021} for feature processing and use `key' features from the 9th ViT layer as our feature representation, and use logarithmic spatial binning of features to aggregate local context at each ViT patch location. 
Furthermore, the attention maps in the ViT provide a reasonable foreground segmentation mask.
As such, when computing cyclical distances, we assign infinite distance to any point which lands off the foreground at any stage in the reference-target image cycle (\cref{sec:semantic_correspond}), to ensure that all correspondences are on the objects of interest.
We refer to the supplementary for further implementation details on our method and baselines.

\subsection{Main Results}

\begin{table}[!htb]
    \small
    \addtolength{\tabcolsep}{2pt}
    \centering
    \caption{We report Median Error and Accuracy at 30º, 15º averaged across all 20 categories. We also report Accuracy at 30º broken down by class for an illustrative subset of categories. We provide full, per category breakdowns in the supplementary.} \label{tab:main_res}
    \resizebox{\linewidth}{!}{
    \begin{tabular}{ccccccccc}
    \toprule
        \multicolumn{1}{l}{} & \multicolumn{3}{c}{All Categories}             & \multicolumn{5}{c}{Per Category (Acc30 $\uparrow$)}                            \\
        \cmidrule(rl){2-4}
        \cmidrule(rl){5-9}
        & Med. Err ($\downarrow$) & Acc30 ($\uparrow$) & Acc15 ($\uparrow$) & Bike & Hydrant & M'cycle & Teddy & Toaster \\
        \midrule
        ICP                  & 111.8  & 3.8  & 0.7  & 3.0   & 6.0   & 1.0   & 1.0  & 7 .0  \\
        SIFT                 & 129.4  & 4.0  & 1.5  & 3.0   & 11.0  & 1.0   & 1.0  & 0.0   \\
        SWaV                 & 123.1  & 7.5  & 3.3  & 13.0  & 9.0   & 8.0   & 5.0  & 7.0   \\
        ICP+BV               & 109.3  & 5.4  & 1.2  & 6.0   & 5.0   & 8.0   & 5.0  & 5.0   \\
        PoseContrast         & 111.5  & 6.9  & 1.1  & 2.0   & 4.0   & 13.0  & 4.0  & 12.0  \\
        \midrule
        Ours (K=30)          & 60.2 & 43.5 & 29.0 & 63.0  & 24.0 & 80.0 & 33.0  & 39.0 \\
        Ours (K=50) & \textbf{53.8} & \textbf{46.3} & \textbf{31.1} & \textbf{71.0} & \textbf{26.0} & \textbf{82.0} & \textbf{39.0} & \textbf{42.0} \\
        \bottomrule
    \end{tabular}
    }
\end{table}
We report results averaged over the 20 considered categories in CO3D in the leftmost columns of \cref{tab:main_res}.
We first highlight that the the baselines show poor performance across the reported metrics.
ICP and SIFT perform most poorly, which we attribute to them being designed for within-instance matching. 
Alignment with the SWaV features, which contain more semantic information, fares slightly better, though still only reports a 7.5\% accuracy at 30º.
Surprisingly, we also found PoseContrast to give low accuracies in our setting.
At first glance, this could simply be an artefact of different canonical poses -- between those inferred by the model, and those imposed by the CO3D labels.
However, we note that we \textit{take the difference} between the reference and target poses as our pose prediction, which cancels any constant-offset artefacts in the canonical pose. 

Meanwhile, our method shows substantial improvements over all implemented baselines. 
Our method reports less than half the Median Error aggregated over all categories, and further demonstrates a \textit{six-fold increase} in accuracy at 30º. We also note that this improvement cannot solely be attributed to the scale of DINO's ImageNet pre-training: the SWaV-based baseline also uses self-supervised features trained on ImageNet \cite{caron2020unsupervised}, and PoseContrast is initialised with MoCo-v2 \cite{Moco2020} weights, again from self-supervision on ImageNet.

We find that performance varies substantially according to the specific geometries and appearances of individual categories.
As such, in the rightmost columns of \cref{tab:main_res}, we show per-category results for an illustrative subset of the selected classes in CO3D.
We find that textured objects, which induce high quality correspondences, exhibit better results (e.g `Bike' and `Motorcycle').
Meanwhile, objects with large un-textured regions (e.g `Toaster') proved more challenging. 

The results for `Hydrant' are illustrative of a challenging case. 
In principle, a hydrant has a clearly defined canonical frame, with faucets appearing on only three of its four `faces' (see \cref{fig:examples}).
However, if the model fails to identify all three faucets as salient keypoints for correspondence, the object displays a high degree of rotational symmetry. 
In this case, SIFT, which focuses exclusively on appearance (i.e it does not learn semantics), performs higher than its average, as the hydrant faucets are consistently among the most textured regions on the object.
Meanwhile, our method, which focuses more on semantics, performs worse than its own average on this category.

\subsection{Making use of multiple views} \label{sec:exps_multiview}

\paragraph{The number of target views}: A critical component of our setting is the availability of multiple views of the target object.
We argue that this is important for the computation of zero-shot pose offset between two object instances, as a single image of the target object may not contain any semantic parts in common with the reference image.
An important factor, therefore, is the number of images available in the target sequence.
In principle, if one had infinite views of the target sequence, and camera transformations between each view, the pose estimation problem collapses to that of finding the best view -- a retrieval-based solution.
However, we note that this is unrealistic.
Firstly, running inference on a set of target views is expensive, with the computational burden generally scaling linearly with the number of views.
Secondly, collecting and storing an arbitrarily large number of views is also expensive. 
Finally, the number of views required to densely and uniformly sample viewpoints of an object is very high, as it requires combinatorially sampling with respect to three rotation parameters. 

In this work we experiment with the realistic setting of a `handful' of views of the target object. 
In \cref{tab:ablation_n_tgt}, we experiment with varying $N$ in $\{1, 3, 5\}$ instances in the target sequence.
In the bottom three rows, we show the performance of our full method as $N$ is varied and find that, indeed, the performance increases with the number of available views (further results are in supplementary).
However, we find that even from a \textit{single view}, our method can outperform the baselines with access to five views.

\paragraph{Only finding best target view}: We disambiguate the `coarse' and `fine' pose-estimation steps of our method (\cref{sec:best_view} and \cref{sec:pose_computation_with_depth} respectively). 
Specifically, we experiment with our method's performance if we assume the reference image is perfectly aligned with the selected best target view. 
We show these figures as `\textit{Ours-BV}' in the top rows of \cref{tab:ablation_n_tgt}.
It can be seen that this part of our method alone can substantially outperform the strongest baselines. 
However, we also show that the subsequent fine alignment step using the depth information (\cref{sec:pose_computation_with_depth}) provides an important improvement in performance.
For instance, when $N=5$, this component of our method boosts Acc30 from 35.4\% to 46.3\%.

\paragraph{How to pick the best target view}: \cref{tab:ablation_best_view} shows the results of a comparison of different methods for selecting the optimal target view. Our method described in \cref{sec:best_view} is denoted \textit{`CorrespondSim'}. \textit{`GlobalSim'} is a baseline where the target view is selected which maximises the similarity with respect to the ViT's global feature vector. We also try maximising the Intersection-over-Union (IoU) of the foreground masks, as provided by the ViT attention maps, of the reference and target frames (\textit{`SaliencyIoU'}).
Lastly, \textit{`CyclicalDistIoU'} chooses the target view which has maximum IoU between the foreground mask of a target object and a thresholded cyclical distance map to the reference image. The intuition here is to recover a target view where a large proportion of the foreground object pixels have a unique nearest neighbour in the reference image.

\begin{figure}
\centering
\begin{floatrow}
\capbtabbox[0.5\textwidth]{%
  \addtolength{\tabcolsep}{2pt}
    \begin{tabular}
        {cccc}
            \toprule
            Method & Med. Err & Acc30 & Acc15 \\ \hline
            Ours-BV (N=1)  & 92.8  & 12.6  &  3.7  \\ 
            Ours-BV (N=3)  & 69.5  & 26.1  &  8.0  \\ 
            Ours-BV (N=5)  & 61.1  & 35.4  &  10.6 \\ \hline
            Ours (N=1)     & 97.3  & 23.8  &  13.4  \\ 
            Ours (N=3)     & 63.9  & 38.9  &  26.2 \\ 
            Ours (N=5)     & \textbf{53.8} & \textbf{46.3} & \textbf{31.1} \\ 
            \bottomrule
        \end{tabular}
}{%
  \caption{We experiment with varying numbers of images available in the target sequence ($N$). Even with only one view, our method substantially outperforms existing baselines with access to multiple views. We further show the utility of pose alignment from the best view (`Ours') over simply choosing the best view with our method (`Ours-BV').}%
  \label{tab:ablation_n_tgt}
}
\capbtabbox[0.5\textwidth]{%
  \begin{tabular}
    {cccc}
        \toprule
        Method & Med. Err & Acc30 & Acc15 \\ \hline
        CyclicalDistIOU    & 94.7  &  22.3  & 13.1   \\ 
        GlobalSim          & 71.9  &  36.8  & 23.7  \\ 
        SaliencyIOU        & 87.0  &  29.1  & 18.1  \\ 
        CorrespondSim      & \textbf{53.8} & \textbf{46.3} & \textbf{31.1} \\ 
        \bottomrule
    \end{tabular}
}{%
  \caption{We ablate different methods for selecting the best view from the target sequence, from which we perform our final pose computation. Compared to a other intuitive options for this task, we demonstrate the importance of our proposed best view selection pipeline for downstream performance.}%
  \label{tab:ablation_best_view}
}
\end{floatrow}
\end{figure}

\subsection{Qualitative Results}

\begin{figure}[h]
\centering
  \includegraphics[width=0.98\columnwidth]{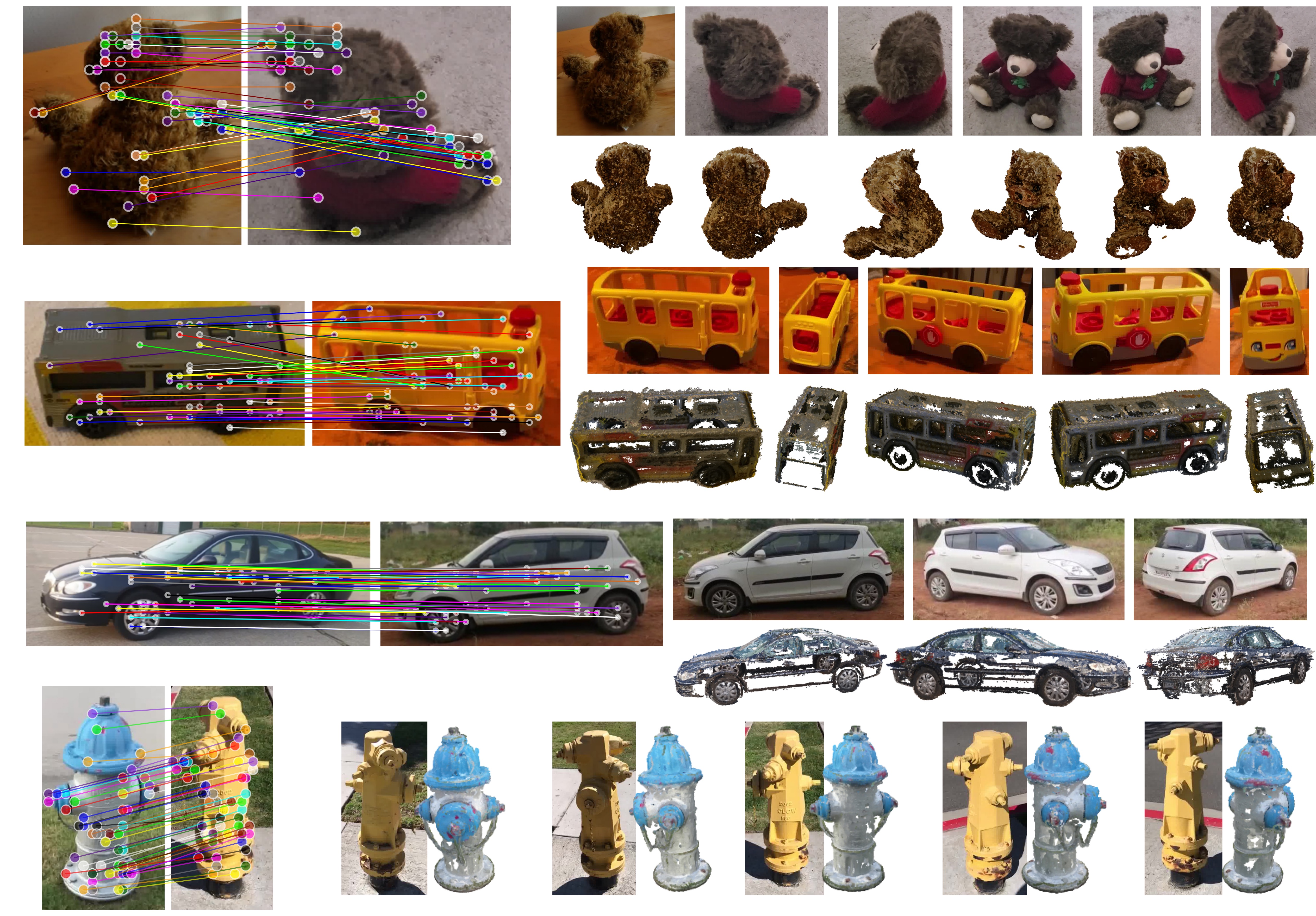}
  \caption{Example results for the categories \textbf{Teddybear}, \textbf{Toybus}, \textbf{Car}, \textbf{Hydrant}. Depicted are the correspondences found between the reference image and the best-matching frame from the target sequence found following \cref{sec:best_view}. To the right, the estimated pose resulting from these correspondences is shown as an alignment between the reference object (shown as a rendered point cloud) and the target sequence. 
  \textbf{Hydrant} depicts a failure mode --- while the result looks visually satisfying, near rotational symmetry (about the vertical axis) leads to poor alignment.}
  \label{fig:examples}
\end{figure}

In \cref{fig:examples} we provide qualitative alignment results for four object categories, including for `Hydrant', which we include as a failure mode.
The images show the reference image and the best view from the target sequence, along with the semantic correspondences discovered between them. 
We further show the point cloud for the reference image aligned with the target sequence using our method. 
Specifically, we first compute a relative pose offset between the reference image and the best target view, and then propagate this pose offset using camera extrinsics to the other views in the target sequence. 
Finally, we highlight the practical utility of this setting.
Consider a household robot wishing to tidy away a `Teddybear' (top row) into a canonical pose (defined by a reference image).
Using this method, the agent is able to view the toy from a number of angles (in the target sequence), align the reference image to an appropriate view, and thus understand the pose of the toy \textit{from any other angle}.

\section{Conclusion} \label{sec:conclusions} 

\paragraph{Consideration of limitations:} We have have proposed a model which substantially outperforms existing applicable baselines for the task of zero-shot category-level pose detection.
However, absolute accuracies remain low and we suggest fall far short of human capabilities. 
Firstly, our performance across the considered classes is 46.3\% Acc30 with 5 views available. We imagine these accuracies to be substantially lower than the a human baseline for this task.
Secondly, though single view novel category alignment is highly challenging for machines, humans are capable of generalising highly abstract concepts to new categories, and thus would likely be able to perform reasonably in a single view setting. 

\vspace{-0.3em}
\paragraph{Final Remarks:} In this paper we have proposed a highly challenging (but realistic) setting for object pose estimation, which is a critical component in most 3D vision pipelines. 
In our proposed setting, a model is required to align two instances of an object category without having any pose-labelled data for training.
We further re-purpose the recently released CO3D dataset and devise a test setting which reasonably resembles the one encountered by a real-world embodied agent.
Our setting presents a complex problem which requires both semantic and geometric understanding, and we show that existing baselines perform poorly on this task.
We further propose a novel method for zero-shot, category-level pose estimation based on semantic correspondences and show it can offer a six-fold increase in Acc30 on our proposed evaluation setting.
We hope that this work will serve as a spring-board to foster future research in this important direction.

\section{Acknowledgment}
The authors gratefully acknowledge the use of the University of Oxford Advanced Research Computing (ARC) facility \url{http://dx.doi.org/10.5281/zenodo.22558}. Sagar Vaze is funded by a Facebook Research Scholarship. We thank Dylan Campbell, Nived Chebrolu and Matias Mattamala for many useful discussions.

\clearpage
\bibliographystyle{splncs04}
\bibliography{pose.bib}

\begin{thebibliography}{10}
\providecommand{\url}[1]{\texttt{#1}}
\providecommand{\urlprefix}{URL }
\providecommand{\doi}[1]{https://doi.org/#1}

\bibitem{Aberman2018}
Aberman, K., Liao, J., Shi, M., Lischinski, D., Chen, B., Cohen-Or, D.: {Neural
  best-buddies: Sparse cross-domain correspondence}. ACM Transactions on
  Graphics  (2018)

\bibitem{AhmadyanObjectron2021}
Ahmadyan, A., Zhang, L., Ablavatski, A., Wei, J., Grundmann, M.: {Objectron: A
  Large Scale Dataset of Object-Centric Videos in the Wild with Pose
  Annotations}. In: CVPR (2021)

\bibitem{Akizuki2021}
Akizuki, S.: {ASM-Net : Category-level Pose and Shape Estimation Using
  Parametric Deformation}. In: BMVC (2021)

\bibitem{Amir2021}
Amir, S., Gandelsman, Y., Bagon, S., Dekel, T.: {Deep ViT Features as Dense
  Visual Descriptors}  (2021)

\bibitem{NavierInpaint}
Bertalmio, M., Bertozzi, A., Sapiro, G.: Navier-stokes, fluid dynamics, and
  image and video inpainting. In: CVPR (2001)

\bibitem{caron2020unsupervised}
Caron, M., Misra, I., Mairal, J., Goyal, P., Bojanowski, P., Joulin, A.:
  Unsupervised learning of visual features by contrasting cluster assignments.
  NeurIPS  (2020)

\bibitem{Caron2021}
Caron, M., Touvron, H., Misra, I., J{\'{e}}gou, H., Mairal, J., Bojanowski, P.,
  Joulin, A.: {Emerging Properties in Self-Supervised Vision Transformers}. In:
  ICCV (2021)

\bibitem{ChenCASS2020}
Chen, D., Li, J., Wang, Z., Xu, K.: {Learning canonical shape space for
  category-level 6D object pose and size estimation}. In: CVPR (2020)

\bibitem{ChenSGPA2021}
Chen, K., Dou, Q.: {SGPA: Structure-Guided Prior Adaptation for Category-Level
  6D Object Pose Estimation}. In: ICCV (2021)

\bibitem{ChenFSNet2021}
Chen, W., Jia, X., Chang, H.J., Duan, J., Shen, L., Leonardis, A.: {FS-Net:
  Fast Shape-based Network for Category-Level 6D Object Pose Estimation with
  Decoupled Rotation Mechanism}. In: CVPR (2021)

\bibitem{Moco2020}
Chen, X., Fan, H., Girshick, R.B., He, K.: {Improved Baselines with Momentum
  Contrastive Learning}  (2020), \url{https://arxiv.org/abs/2003.04297}

\bibitem{Chen2020}
Chen, X., Dong, Z., Song, J., Geiger, A., Hilliges, O.: {Category Level Object
  Pose Estimation via Neural Analysis-by-Synthesis}. In: ECCV (2020)

\bibitem{cheng2021improving}
Cheng, X., Lin, H., Wu, X., Yang, F., Shen, D.: Improving video-text retrieval
  by multi-stream corpus alignment and dual softmax loss (2021)

\bibitem{ChoiRedwood2016}
Choi, S., Zhou, Q.Y., Miller, S., Koltun, V.: {A Large Dataset of Object Scans}
   (2016)

\bibitem{Deng2020}
Deng, X., Xiang, Y., Mousavian, A., Eppner, C., Bretl, T., Fox, D.:
  {Self-supervised 6D Object Pose Estimation for Robot Manipulation}. In: ICRA
  (2020)

\bibitem{dosovitskiy2020vit}
Dosovitskiy, A., Beyer, L., Kolesnikov, A., Weissenborn, D., Zhai, X.,
  Unterthiner, T., Dehghani, M., Minderer, M., Heigold, G., Gelly, S.,
  Uszkoreit, J., Houlsby, N.: An image is worth 16x16 words: Transformers for
  image recognition at scale. ICLR  (2021)

\bibitem{Eggert1997}
Eggert, D.W., Lorusso, A., Fisher, R.B.: {Estimating 3-D rigid body
  transformations: A comparison of four major algorithms}. Machine Vision and
  Applications  (1997)

\bibitem{Banani2020}
{El Banani}, M., Corso, J.J., Fouhey, D.F.: {Novel object viewpoint estimation
  through reconstruction alignment}. In: CVPR (2020)

\bibitem{Florence2018}
Florence, P.R., Manuelli, L., Tedrake, R.: {Dense Object Nets: Learning Dense
  Visual Object Descriptors By and For Robotic Manipulation}. In: CoRL (2018)

\bibitem{Goodwin2021}
Goodwin, W., Vaze, S., Havoutis, I., Posner, I.: {Semantically Grounded Object
  Matching for Robust Robotic Scene Rearrangement}. In: ICRA (2022)

\bibitem{Grabner2018}
Grabner, A., Roth, P.M., Lepetit, V.: {3D Pose Estimation and 3D Model
  Retrieval for Objects in the Wild}. In: CVPR (2018)

\bibitem{GuptaPose2015}
Gupta, S., Arbel{\'{a}}ez, P., Girshick, R., Malik, J.: {Inferring 3D Object
  Pose in RGB-D Images}  (2015)

\bibitem{Hinterstoisser2013}
Hinterstoisser, S., Lepetit, V., Ilic, S., Holzer, S., Bradski, G., Konolige,
  K., Navab, N.: {Model based training, detection and pose estimation of
  texture-less 3D objects in heavily cluttered scenes}. In: Lecture Notes in
  Computer Science (2013)

\bibitem{Huynh09}
Huynh, D.Q.: Metrics for 3d rotations: Comparison and analysis. J. Math.
  Imaging Vis.  (2009)

\bibitem{Kanezaki2018}
Kanezaki, A., Matsushita, Y., Nishida, Y.: {RotationNet: Joint Object
  Categorization and Pose Estimation Using Multiviews from Unsupervised
  Viewpoints}. In: CVPR (2018)

\bibitem{Kundu2019}
Kundu, J.N., Rahul, M.V., Ganeshan, A., Babu, R.V.: {Object pose estimation
  from monocular image using multi-view keypoint correspondence}. In: ECCV
  (2019)

\bibitem{Lee2019SFNET}
Lee, J., Kim, D., Ponce, J., Ham, B.: {SFNET: Learning object-aware semantic
  correspondence}. In: CVPR (2019)

\bibitem{Li2021SE3}
Li, X., Weng, Y., Yi, L., Guibas, L., Abbott, A.L., Song, S., Wang, H.:
  {Leveraging SE(3) Equivariance for Self-Supervised Category-Level Object Pose
  Estimation}. In: NeurIPS 2021 (2021)

\bibitem{Lin2021Pose}
Lin, Y., Tremblay, J., Tyree, S., Vela, P.A., Birchfield, S.: {Single-stage
  Keypoint-based Category-level Object Pose Estimation from an RGB Image}. In:
  ICRA (2022)

\bibitem{liu2020semantic}
Liu, Y., Zhu, L., Yamada, M., Yang, Y.: Semantic correspondence as an optimal
  transport problem. In: CVPR (2020)

\bibitem{Lowe2004SIFT}
Lowe, D.G.: {Distinctive Image Features from Scale-Invariant Keypoints}.
  International Journal of Computer Vision  (2004)

\bibitem{Manuelli2019}
Manuelli, L., Gao, W., Florence, P., Tedrake, R.: {kPAM: KeyPoint Affordances
  for Category-Level Robotic Manipulation}. In: ISRR (2019)

\bibitem{Pavlakos2017}
Pavlakos, G., Zhou, X., Chan, A., Derpanis, K.G., Daniilidis, K.: {6-DoF Object
  Pose from Semantic Keypoints}. In: ICRA (2017)

\bibitem{Pytorch3d}
Ravi, N., Reizenstein, J., Novotny, D., Gordon, T., Lo, W., Johnson, J.,
  Gkioxari, G.: {Accelerating 3D Deep Learning with PyTorch3D}  (2020),
  \url{https://arxiv.org/abs/2007.08501}

\bibitem{ReizensteinCO3D2021}
Reizenstein, J., Shapovalov, R., Henzler, P., Sbordone, L., Labatut, P.,
  Novotny, D.: {Common Objects in 3D: Large-Scale Learning and Evaluation of
  Real-life 3D Category Reconstruction}. In: ICCV (2021)

\bibitem{Sahin2019}
Sahin, C., Kim, T.K.: {Category-level 6D object pose recovery in depth images}.
  In: ECCV (2019)

\bibitem{Sajnani2022}
Sajnani, R., Poulenard, A., Jain, J., Dua, R., Guibas, L.J., Sridhar, S.:
  {ConDor: Self-Supervised Canonicalization of 3D Pose for Partial Shapes}. In:
  CVPR (2022)

\bibitem{Schoenberger2016}
Schonberger, J.L., Frahm, J.M.: {Structure-from-Motion Revisited}. In: CVPR
  (2016)

\bibitem{Shi2021}
Shi, J., Yang, H., Carlone, L.: {Optimal Pose and Shape Estimation for
  Category-level 3D Object Perception}. In: RSS (2021)

\bibitem{Simeonov2021}
Simeonov, A., Du, Y., Tagliasacchi, A., Tenenbaum, J.B., Rodriguez, A.,
  Agrawal, P., Sitzmann, V.: {Neural Descriptor Fields: SE(3)-Equivariant
  Object Representations for Manipulation}. In: ICRA (2022)

\bibitem{Tian2020}
Tian, M., Ang, M.H., Lee, G.H.: {Shape Prior Deformation for Categorical 6D
  Object Pose and Size Estimation}. In: ECCV (2020)

\bibitem{Tseng2019}
Tseng, H.Y., {De Mello}, S., Tremblay, J., Liu, S., Birchfield, S., Yang, M.H.,
  Kautz, J.: {Few-shot viewpoint estimation}. In: BMVC (2020)

\bibitem{Umeyama1991}
Umeyama, S.: {Least-Squares Estimation of Transformation Parameters Between Two
  Point Patterns}. IEEE PAMI  (1991)

\bibitem{vaze22gcd}
Vaze, S., Han, K., Vedaldi, A., Zisserman, A.: {Generalized Category
  Discovery}. In: CVPR (2022)

\bibitem{Wang2021_NeMo}
Wang, A., Kortylewski, A., Yuille, A.: {NeMo: Neural Mesh Models of Contrastive
  Features for Robust 3D Pose Estimation}. In: ICLR (2021)

\bibitem{Wang2019NOCS}
Wang, H., Sridhar, S., Huang, J., Valentin, J., Song, S., Guibas, L.:
  {Normalized Object Coordinate Space for Category-Level 6D Object Pose and
  Size Estimation}. In: CVPR (2019)

\bibitem{Xiang2016}
Xiang, Y., Kim, W., Chen, W., Ji, J., Choy, C., Su, H., Mottaghi, R., Guibas,
  L., Savarese, S.: {Objectnet3D: A large scale database for 3D object
  recognition}. In: ECCV (2016)

\bibitem{Xiang2014}
Xiang, Y., Mottaghi, R., Savarese, S.: {Beyond PASCAL: A benchmark for 3D
  object detection in the wild}. In: WACV (2014)

\bibitem{Xiang2018}
Xiang, Y., Schmidt, T., Narayanan, V., Fox, D.: {PoseCNN: A Convolutional
  Neural Network for 6D Object Pose Estimation in Cluttered Scenes}. In: RSS
  (2018)

\bibitem{Xiao2021}
Xiao, Y., Du, Y., Marlet, R.: {PoseContrast: Class-Agnostic Object Viewpoint
  Estimation in the Wild with Pose-Aware Contrastive Learning}. In: 3DV (2021)

\bibitem{Xiao2020}
Xiao, Y., Marlet, R.: {Few-Shot Object Detection and Viewpoint Estimation for
  Objects in the Wild}. In: ECCV (2020)

\bibitem{Xiao2019}
Xiao, Y., Qiu, X., Langlois, P.A., Aubry, M., Marlet, R.: {Pose from Shape:
  Deep pose estimation for arbitrary 3D objects}. In: BMVC (2019)

\bibitem{Open3D2018}
Zhou, Q., Park, J., Koltun, V.: Open3d: {A} modern library for 3d data
  processing  (2018), \url{http://arxiv.org/abs/1801.09847}

\bibitem{Zhou2018}
Zhou, X., Karpur, A., Luo, L., Huang, Q.: {StarMap for Category-Agnostic
  Keypoint and Viewpoint Estimation}. In: ECCV (2018)

\end{thebibliography}
\clearpage
\titlerunning{Zero-Shot Category-Level Object Pose Estimation: Supplementary Material}
\authorrunning{W. Goodwin et al.}
\section{Supplementary Material}
\renewcommand{\thefigure}{A.\arabic{figure}}
\setcounter{figure}{0} 
\renewcommand{\thetable}{A.\arabic{table}}
\setcounter{table}{0} 

In this appendix, we first discuss our choice of dataset, followed by our choice of evaluation categories and sequences, and a description of our pose-labelling procedure, and data pre-processing steps. We then present several further experiments and ablations to our method, showing that performance improves further under greater numbers of target views, and the effectiveness of our full method in refining a pose estimation. Results around the number and diversity of correspondences are given, and the approach to the rigid body transform solution and RANSAC is described further and justified. We examine our choice of evaluation metric for the SO(3) component of pose estimation, and explore the effect of near-symmetries on our results in this light. We give further implementation details on several baselines. 

\appendix
\section{CO3D dataset}
\subsection{Choice of dataset}
\begin{table}
\centering
     \resizebox{\linewidth}{!}{
    \begin{tabular}{cccccccc}
    \toprule
        \textbf{Dataset} & \textbf{\# Cat.}  & \textbf{\# Obj./Cat.} & \textbf{\# View/Obj.}  & \textbf{Pose} & \textbf{Extrinsics} & \textbf{Depth} & \textbf{Pcd/Mesh}        \\
        \midrule
        \textbf{Pascal3D} \cite{Xiang2014} & 12 & $\sim$3000 & 1 & Yes & No & No & No \\
        \textbf{ObjectNet3D} \cite{Xiang2016} & 100  & 2019 & 1 & Yes & No & No & No \\
        \textbf{Objectron} \cite{AhmadyanObjectron2021} & 9 & 1621 & 268 & Yes & Yes & No* & No* \\
        \textbf{Redwood} \cite{ChoiRedwood2016} & 320 & $\sim$28 & $\sim$2300 & No & No & Yes & No\textsuperscript{†} \\
        \textbf{REAL275} \cite{Wang2019NOCS} & 6 & 7 & $\sim$950\textsuperscript{‡} & Yes & Yes & Yes & Yes \\
        \textbf{CO3D} \cite{ReizensteinCO3D2021} & 51 & $\sim$380 & $\sim$79 & No & Yes & Yes & Yes \\
        \bottomrule
    \end{tabular}
    }
\caption{A comparison of multi-view category-level datasets (\textbf{\# Cat.} = number of categories, \textbf{\# Obj./Cat} = average number of distinct object instances per category, \textbf{\# View/Obj.} = average number of views of each distinct instance). We find that \textbf{CO3D} is the only dataset that offers a large number of categories, with diversity within the category, alongside multiple views and depth information for each object. *Depth and point cloud information in \textbf{Objectron} is only available via the highly sparse points used in the SfM process. \textsuperscript{†}The Redwood dataset provides high quality mesh reconstructions for just 398 object instances, from a subset of only 9 categories. \textsuperscript{‡}The combined train/val/test splits of \textbf{REAL275} contain 8,000 frames, each with at least 5 objects present. With 42 object instances, this gives $\sim$950 appearances per instance.}%
\label{tab:datasets}
\end{table}

A comparison of several multi-category, multi-instance datasets is given in \cref{tab:datasets}. Several existing canonical category-level pose datasets are not appropriate for our method as they do not include depth information \cite{Xiang2014, Xiang2016}, or only have extremely sparse depth \cite{AhmadyanObjectron2021}. The \textbf{Redwood} dataset \cite{ChoiRedwood2016} contains a good diversity of object categories and instances, with many views per object and ground truth depth maps, but structure-from-motion (SfM) is only run on a small subset of categories and sequences, so very few sequences have camera extrinsics, required to evaluate the multiple target view version of our method. The \textbf{REAL275} dataset \cite{Wang2019NOCS}, being motivated in the same embodied settings as the present work, has the appropriate depth and extrinsic information. However, the dataset contains only 6 categories and a small number of instances (7 per category). The present work considers a zero-shot approach to category-level pose, and a strong quantitative and qualitative evaluation of this method requires a large diversity of object categories. \textbf{CO3D} \cite{ReizensteinCO3D2021} provides this, with 51 object categories, each containing a large variety of instances, with depth and camera extrinsic information. While unlike most of the other datasets considered in \cref{tab:datasets}, CO3D does not contain labelled category-level pose, we find that we are able to label sufficient sequences ourselves to reach robust quantitative evaluation of our methods and baselines (\cref{sec:co3d_labelling}). As our method is fully unsupervised, we do not require a large labelled dataset for training: a sufficient test set is all that is needed.

\subsection{Choice of evaluation categories \& sequences} \label{sec:cat_choice}
The CO3D dataset contains hundreds of sequences for each of 51 distinct object categories. In this work, our quantitative evaluation is performed on a subset of 20 of these categories. We \textbf{\textit{exclude}} categories based on the following criteria:
\begin{itemize}
    \item Categories for which the object has one or more axes of infinite rotational symmetry. 16 categories (\textit{apple, ball, baseball bat, bottle, bowl, broccoli, cake, carrot, cup, donut, frisbee, orange, pizza, umbrella, vase, wineglass}).
    \item Categories for which the object has more than one rotational symmetry. 6 categories (\textit{bench, hot dog, kite, parking meter, skateboard, suitcase}).
    \item Categories for which an insufficient number of sequences ($<10$) have high-quality point clouds and camera viewpoints. 6 categories (\textit{banana, cellphone, couch, microwave, stop sign, TV}).
    \item Categories for which between-instance shape difference made labelling challenging or fundamentally ambiguous. 3 categories (\textit{baseball glove, plant, sandwich}).
\end{itemize}
This leaves 20 categories, as shown in \cref{fig:err_distrib}. Some included categories were still `marginal' under these criteria, for instance \textit{handbag}, where there was a 180º rotational symmetry for most instances. Here, the labelling convention was to, where possible, disambiguate pose labels by which side of the handbag the handle fell onto. Nonetheless, categories such as \textit{handbag} and \textit{toaster} elicited bi-modal predictions from our method, reflecting these ambiguities, as shown in \cref{fig:err_distrib}.

We further select a subset of sequences for labelling (\cref{sec:co3d_labelling}) from each of these 20 categories. CO3D provides predicted quality scores for camera viewpoints and point clouds reconstructed by the COLMAP structure-from-motion (SfM) processes \cite{ReizensteinCO3D2021}. Each category has an average of 356 sequences (distinct object instances), ranging from 21 for \textit{parking meter} to 860 for \textit{backpack}. We choose to consider all sequences that have a viewpoint quality score of more than $1.25$, and a point cloud quality of greater than $0.3$. On average, this is the top $16\%$ of sequences within a category, and returns a median of 36 valid sequences per category. For our chosen categories (\cref{sec:cat_choice}), we choose to label the top 10 sequences based on point cloud scores with category-level pose.

\subsection{Labelling pose for evaluation} \label{sec:co3d_labelling}
\begin{figure}[tbh]
\centering
  \includegraphics[width=0.98\columnwidth]{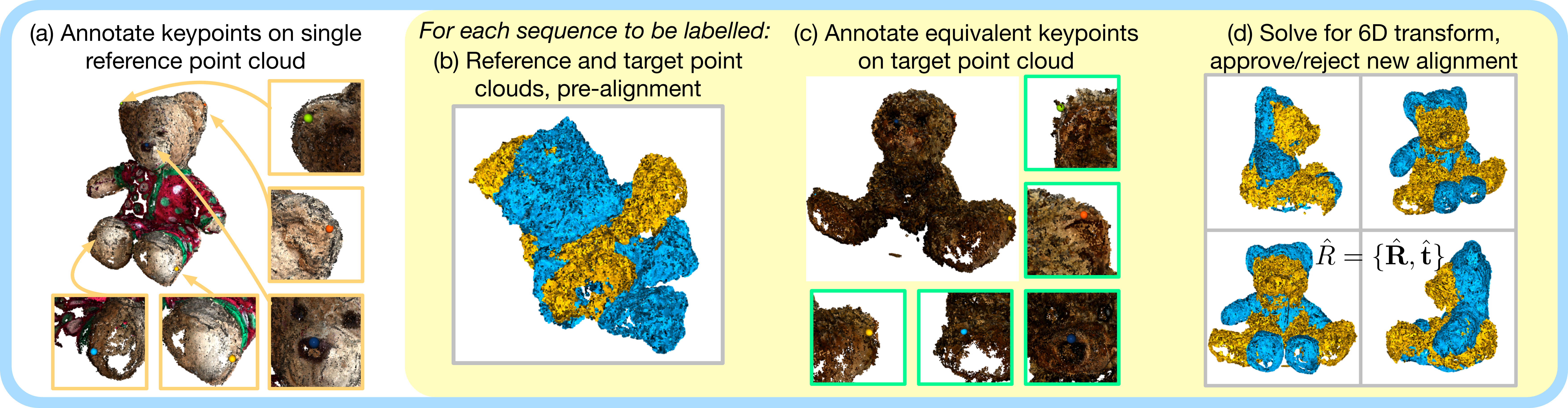}
  \caption{The process used in this work to generate category-level pose labels for the CO3D dataset, in the presence of large between-instance shape and appearance shift. Our interface uses Open3D \cite{Open3D2018} for annotation and visualisation.}
  \label{fig:labelling}
\end{figure}

The per-frame camera extrinsics in CO3D are given relative to the first frame in each sequence.
Thus, the camera extrinsic positions do not relate the SE(3) poses of objects within a category with respect to any category-level canonical pose. Indeed, this is noted by the dataset's authors \cite{ReizensteinCO3D2021} as a limitation of using the dataset to learn category-level object representations. To overcome this and enable quantitative evaluation, we design a labelling interface that leverages the sequence point clouds for fast and intuitive category-level pose alignment. The process is depicted in \cref{fig:labelling}. For each category, we choose the sequence with the highest point cloud quality score to be the reference object. Four or more semantically salient keypoints that are deemed likely to exist, in a spatially consistent manner, across all instances in the category are selected interactively on this point cloud, using the interface. Subsequently, the labeller is presented with the other candidate objects in turn, and selects the equivalent points in the same order. Umeyama's method is then used to solve for the rigid body transform and uniform scaling, given annotated keypoint correspondences \cite{Umeyama1991}. The labeller is then presented with the reference point cloud, overlaid with the transformed target point cloud, both coloured uniformly for clarity, and can inspect the quality of the alignment. If it is adequate, the transform is accepted, and the rigid body parameters $\hat{T}=\left(\hat{\mathbf{R}}, \hat{\mathbf{t}}\right)$ saved as a pose label relative to the target sequence. This provides labels of pose offsets at the point cloud level, which is in world coordinate space. Every frame in a sequence is related to the world coordinates via the predicted camera extrinsics. Further, every sequence will have a relative pose against the reference sequence's point cloud. Using this, a ground-truth relative pose in the camera frame, which is what our method predicts, can be constructed for any two frames $i$ and $j$ from any two sequences $a$ and $b$ as:
\begin{equation}
    \mathbf{T}_{a_{i}b_{j}} = (\mathbf{T}^{\text{cam}}_{a_{i}})^{-1} \circ \mathbf{T}_{0a}^{-1} \circ \mathbf{T}_{0b} \circ \mathbf{T}^{\text{cam}}_{b_{j}}
\end{equation}
Where $\mathbf{T}$ denotes a $4\times4$ homogeneous transform matrix composed from rotation $\mathbf{R}$ and translation $\mathbf{t}$, and $\mathbf{T}_{0a}$, $\mathbf{T}_{0b}$ are the transforms from reference to target object point clouds as computed in our labelling procedure, and $\mathbf{T}^{\text{cam}}_{a_{i}}$, $\mathbf{T}^{\text{cam}}_{b_{j}}$ are the camera extrinsics (world to view transforms) from the SfM procedure in CO3D. $\circ$ denotes function composition - as these functions are transformation matrices, the resultant transform is $\mathbf{T}^{\text{cam}}_{b_{j}} \mathbf{T}_{0b} \mathbf{T}_{0a}^{-1} (\mathbf{T}^{\text{cam}}_{a_{i}})^{-1}$.

\subsection{Data processing}
\subsubsection{Depth completion}
CO3D uses crowd-sourced video, with the original data coming from RGB cameras before structure-from-motion is extracted by COLMAP \cite{Schoenberger2016}. CO3D chooses to scale all point clouds to have unit standard deviation averaged across 3 world coordinate axes, which then fixes the camera intrinsics and depth maps to be relative to this world coordinate scheme. For our purposes, this scale ambiguity is acceptable - we can nonetheless evaluate SE(3) pose predictions, for which the rotation component is independent of scale, and for which the translation component will be affected but still has a well-posed and recoverable ground truth. 

On the other hand, the depth maps in CO3D are estimates from COLMAP's multi-view stereo (MVS) algorithm, and are incomplete. Our method requires accurate depth to project the discovered semantic correspondences into 3D space, enabling a solution for the rigid body transform between object instances (\cref{sec:pose_computation_with_depth}).
One approach would be to disregard those correspondences that land on an area with unknown depth. However, as the correspondences are found at the ViT patch level ($8 \times 8$ pixels, see \cref{sec:semantic_correspond}), we found a small number of missing areas in the per-pixel depth maps led to throwing away a disproportionate amount of correspondences.
Instead, we use a fast in-painting method based on the Navier-Stokes equations \cite{NavierInpaint}, implemented in OpenCV, to fill missing values.

\subsubsection{Object crops}
CO3D uses a supervised segmentation network to produce probabilistic mask labels for every frame. We threshold these and pad the result by 10\% to give a region of interest for the objects. We use this to crop the depth maps and RGB images when evaluating our method. However, we do not use these masks further within our method.

\section{Further experiments}

\subsection{Number of target views} \label{sec:num_views}
\begin{figure}
\centering
\begin{floatrow}
\capbtabbox[0.6\textwidth]{%
  \addtolength{\tabcolsep}{2pt}
    \begin{tabular}{ccccc}
    \toprule
        \multicolumn{1}{l}{} & \multicolumn{2}{c}{Best view only}   & \multicolumn{2}{c}{Full method}      \\
        \cmidrule(rl){2-3}
        \cmidrule(rl){4-5}
        Target views & Acc30 $\uparrow$ & Acc15 $\uparrow$ & Acc30 & Acc15   \\
        \midrule
        1   &  12.6 & 3.7  & \textbf{23.8} & \textbf{13.4}    \\
        3   &  26.1 & 8.0  & \textbf{38.9} & \textbf{26.2}   \\
        5   &  35.4 & 10.6 & \textbf{46.3} & \textbf{31.1}   \\
        10  &  45.0 & 16.2 & \textbf{52.5} & \textbf{38.2}   \\
        20  &  47.0 & 18.6 & \textbf{52.1} & \textbf{38.3}   \\
        \bottomrule
    \end{tabular}
}{%
  \caption{(Acc30, Acc15 = percentage of predictions with a geodesic error of less than 30º, 15º.). An extension of the comparison in sec. 5.5 of the effect of increasing the number of available target views, and the improvement of the full method including solving for a rigid body transformation, over just taking the best view as a pose prediction.}%
  \label{tab:ablation_n_tgt_full}
}
\ffigbox[0.4\textwidth]{
\centering
  \includegraphics[width=0.38\textwidth]{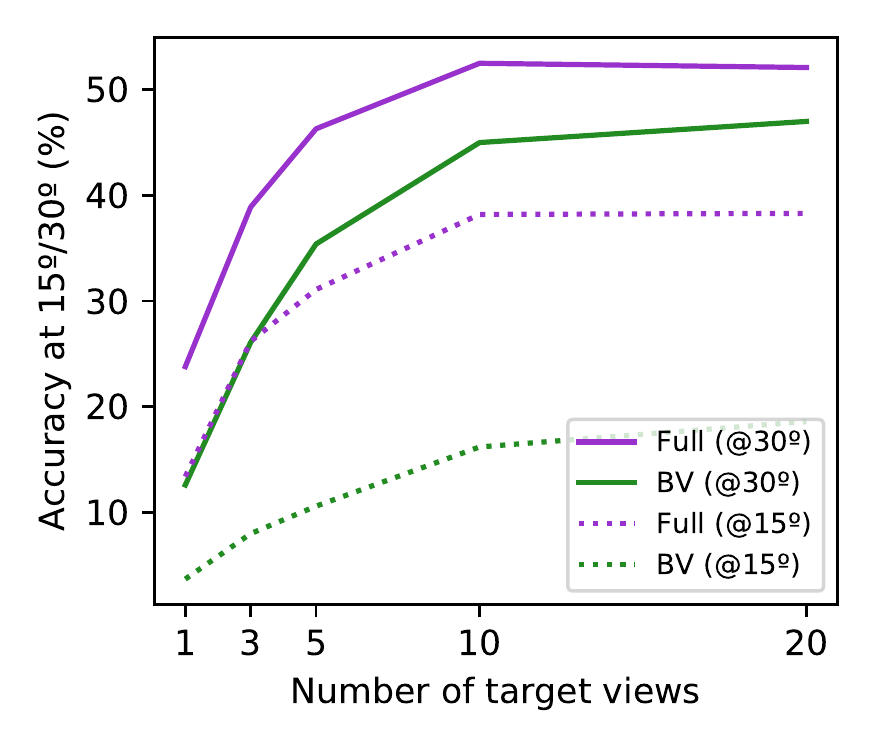}
  \label{fig:bestview}
}{\caption{`Full' method (purple) vs `BV' (best view) only (green). As the number of target views increases, both accuracy metrics improve, though exhibit diminishing returns. The full method leads the best-view ablation throughout, especially in Acc15.}}
\end{floatrow}
\end{figure}

In the main paper, we show that the number of available target views is an important parameter in our method, demonstrating that as we increase from 1 view to 3 and 5 views, pose estimation performance improves. Here, we include two further results, in which 10 and 20 target views are available, to investigate whether these effects continue to scale. We also present the results achieved by taking the coarse pose estimate given by our best view selection method, without further refinement from the rigid body transform component. The results are shown in Tab.~\ref{tab:ablation_n_tgt_full} and Fig. A.2. Clearly, increasing the number of target views available has a positive effect on performance, though in an embodied setting this would come at the cost of the time to explore and image multiple views. While it can be seen that by doubling from 5 to 10 target views improves the Acc30 by over 6\%, we chose to report only the figures for the small number of views (1, 3, 5) in the main text, to reflect such a practical use case. It can also be seen - as already noted in \cref{sec:exps_multiview} - that the full method, including the rigid body transform computed leveraging the semantic correspondences, outperforms the baseline of simply taking the `best' view as predicted by our method's first stage. This continues to hold in the regimes with 10 and 20 target views. Finally, inspecting Fig. A.2 makes it clear that while the full method benefits Acc30, its effect is most marked in improving Acc15 over the performance of taking the best view. This is in line with intuition, which is that the rigid body solution provides fine-tuning on top of a coarse initial estimate (see \cref{sec:exps_multiview}).

\subsection{Design of correspondence method} \label{sec:corresp_method}
\begin{table}[!htb]
\centering
\begin{tabular}
{cccc}
    \toprule
    Method & Med. Err ($\downarrow$) & Acc30 ($\uparrow$) & Acc15 ($\uparrow$) \\ \hline
    Optimal Transport \cite{liu2020semantic}                 & 54.6  &  44.8  & 29.1  \\ 
    Dual Softmax \cite{cheng2021improving}              & 54.0 &  44.5 &  30.1 \\ 
    Cyclical Distances (Ours)             & \textbf{53.8} & \textbf{46.3} & \textbf{31.1} \\ 
    \bottomrule
\end{tabular}
\caption{We ablate our design of correspondence method, which is based on building a cyclical distance map (see Sec. 4.1). Here, we report results of pose estimation using an equivalent map arising from running optimal transport \cite{liu2020semantic} and dual-softmax \cite{cheng2021improving} on top of the raw feature similarity matrix.}%
\label{tab:corresp_method}
\end{table}

Our method builds a cyclical distance map to extract semantic correspondences between two images. 
Here, we experiment with alternate methods of identifying the corresponding locations.
Specifically, our problem is a special case of the general machine learning problem of identifying matches between two sets of features (i.e the spatial features in the target and reference images). 
As such, given a matrix of feature distances between the two normalised feature maps, $D \in \{-1...1\}^{H' \times W' \times H' \times W'}$, we experiment with two additional methods for selecting $K$ entries to serve as correspondences. 
In \cref{tab:corresp_method}, we experiment both with a standard optimal transport solution \cite{liu2020semantic} as well as a recent dual-softmax method \cite{cheng2021improving}, and find our `cyclical distance' design choice is the most suitable for this task.

\subsection{Number and diversity of correspondences}

\begin{figure}
\centering
\begin{tabular}{ccccc}
\toprule
    \multicolumn{1}{l}{} & \multicolumn{2}{c}{Without K-means}   & \multicolumn{2}{c}{With K-means}      \\
    \cmidrule(rl){2-3}
    \cmidrule(rl){4-5}
    \# Correspondences & Acc30 ($\uparrow$)  & Acc15 ($\uparrow$) & Acc30  ($\uparrow$) & Acc15 ($\uparrow$) \\
    \midrule
    30           &  38.7 &  25.2 &  43.5 &  29.0   \\
    50           &  42.0 &  28.6 &  46.3 & \textbf{31.1}   \\
    70           &  43.4 &  30.4 &  \textbf{47.1} & 30.8   \\
    \bottomrule
\end{tabular}
\caption{Comparison of results over 20 categories as the number of correspondences is varied, and when K-means clustering is used to return a set of correspondences that are maximally distinct in descriptor space.}%
\label{tab:kmeans_ncorr}
\end{figure}

In \cref{sec:semantic_correspond}, we describe our approach to guarantee the return of a desired number of correspondences through the introduction of the concept of the `cyclical distance' induced by following a chain of descriptor nearest neighbours from reference image, to target, and back to the reference image. We keep the top-K correspondences under this metric for our method. In some cases, however, there can be a particular region of the two objects that gives rise to a large portion of the top-K correspondences. This can in turn lead to less appropriate pose estimates from the rigid body transform solution (see \cref{sec:rbt}), as a transform can produce this cluster of points and give a large number of inliers for RANSAC, while not aligning the object's in a satisfactory global way. To address this bias, we seek to augment the choice of the top-K correspondences to encourage spatial and semantic diversity. Inspired by \cite{Amir2021}, we employ k-means clustering in descriptor space. We sample the top-$2K$ correspondences under the cyclical distance measure, then seek to produce $K$ clusters. We return a correspondence from each cluster, choosing the one that has the highest ViT salience in the reference image. The effect of this K-means step, and the impact of using differing numbers of correspondences, is shown in \cref{tab:kmeans_ncorr}. We find that k-means clustering improves performance, and use this throughout the other experiments in this paper. We find that using 50 correspondences in our method is sufficient for a trade-off between run-time, correspondence quality, and pose prediction error. 

\subsection{Rigid body transform solution} \label{sec:rbt}
\subsubsection{Algorithm choice}
In our method, given a number of corresponding points in 3D space, we solve for the rigid body transform that minimises the residual errors between the points of the target object, and the transformed points of the reference object. There are a number of solutions to this problem, with some based on quaternions, and some on rotation matrices and the singular value decomposition. A comparison of four approaches is given in \cite{Eggert1997}. We choose to use Umeyama's method \cite{Umeyama1991}, as it allows for simultaneously solving for both the 6D rigid body transform, as well as a uniform scaling parameter. It is also robust under large amounts of noise, while other methods can return reflections rather than true rotations as a degenerate solution \cite{Eggert1997}. 
\subsubsection{RANSAC parameters} 
We performed light tuning of the RANSAC parameters by considering only the teddybear category. Two parameters are important: the maximum number of trials, and the inlier threshold. As the point clouds in CO3D are only recovered up to a scale, the authors choose the convention of scaling them to have a unit standard deviation averaged across the three world axes. This makes the choice of a single inlier threshold to be used across all categories possible. In our experiments, we choose 0.2 as this threshold, which in the context of the rigid body transform solution means that any point that, under the recovered transform, is less than a 0.2 Euclidian distance away from its corresponding point, is considered an inlier. We note that in a real-world setting, the average size of objects is very different between categories, and so it would likely be necessary to use an adaptive threshold on the inlier distance. As we estimate accurate object foreground masks DINO ViT saliency maps \cite{Caron2021}, we can quite accurately estimate object size for both reference and target objects, which could be use to choose this parameter in practice. 

The second important parameter for RANSAC is the number of trials that are run. We chose to limit this to keep inference to a few seconds, and use 1,000 trials for all categories. With 5 target views, this gives the $46.3\%$ Acc30 reported in the main paper. Using 500 trials, this drops to $45.8\%$, and using 2,000 trials, it rises to $46.6\%$.

Finally, we sample 4 correspondences within every RANSAC trial to compute the rigid body transform. Solutions to this problem can suffer from degeneracy with only 3 points \cite{Eggert1997}.

\subsection{Depth-free methods}
\begin{figure}
\centering
\begin{tabular}
{cccc}
    \toprule
     & Med. Err ($\downarrow$) & Acc30 ($\uparrow$) & Acc15 ($\uparrow$) \\ \hline
    Ours ($K=50$, No Depth)            & 81.7 & 23.6 & 6.8 \\
    Ours-BV ($K=50$, With Depth)   & 61.1  & 35.4  &  10.6 \\ 
    Ours ($K=50$, With Depth)            & \textbf{53.8} & \textbf{46.3} & \textbf{31.1} \\
    \bottomrule
\end{tabular}
\caption{We investigate the performance of our method in a depth-free setting, finding that allowing access to depth substantially improves pose estimation performance. `Ours-BV' indicates simply assuming that the best target view (recovered with our method) is perfectly aligned with the reference.}%
\label{tab:depth_free}
\end{figure}

We permit depth maps in our setting as we believe them to be readily available in many practical scenarios, whether through SfM (as in our experiments with CO3D), depth cameras, or stereo. However, here we include a depth-free algorithm where, following `best view’ retrieval with our method, we estimate the essential matrix between this and the reference, and extract pose from this. We show the results in \cref{tab:depth_free}, finding that this depth-free method is not as robust as our depth-based method, for the fine-grained alignment task required after best view retrieval (though it still outperforms the baselines). Specifically, we find pose prediction through essential matrix estimation to be worse than simply predicting an identity transform on top of the best-view. We suggest further investigation is warranted in depth-free variants of our setting.

\subsection{Analysis of results}
\subsubsection{Choice of evaluation metrics}
It has long been noted that when reporting pose estimation errors and accuracies, results can be skewed by the presence of rotationally symmetric objects, where a `reasonable' pose estimate can nonetheless be assigned a very high geodesic error (e.g. a toaster that is predicted to have an azimuth angle of 180º rather than 0º --- both settings would have very similar appearance). For this reason, some works that assume access to object CAD models or point clouds relax the evaluation of pose estimation. For instance, \cite{Hinterstoisser2013} propose the closest point distance metric for symmetric objects, which naturally accounts for symmetries by summing the distances between all points on an object under the predicted pose, and the \textit{closest} points to these on the reference object under the ground-truth pose. 

In this work, we use accuracy (at 15º, 30º) and median error metrics, as is conventional across much of the pose estimation literature. Our reasons for this are twofold. Firstly, cross-instance shape gap makes closest point distance metrics, used in the single-instance literature to handle symmetry, ill-posed. 
A `perfect' relative pose prediction between two object instances would nonetheless carry a non-zero distance due to shape differences. Second, the choice of whether or not to use the closest point distance is made based on whether an object has a rotational symmetry or not \cite{Hinterstoisser2013}. In the zero-shot setting, this cannot be known either a-priori or at test time. Our metrics are thus sensitive to symmetries, but the most appropriate choice for category-level pose estimation. To reduce the impact of symmetries in skewing the reported results, we do not consider object categories with infinite rotational symmetry (see \cref{sec:cat_choice}).

\subsubsection{Impact of near rotational symmetry on results}
\begin{figure}[tb]
\centering
  \includegraphics[width=\columnwidth]{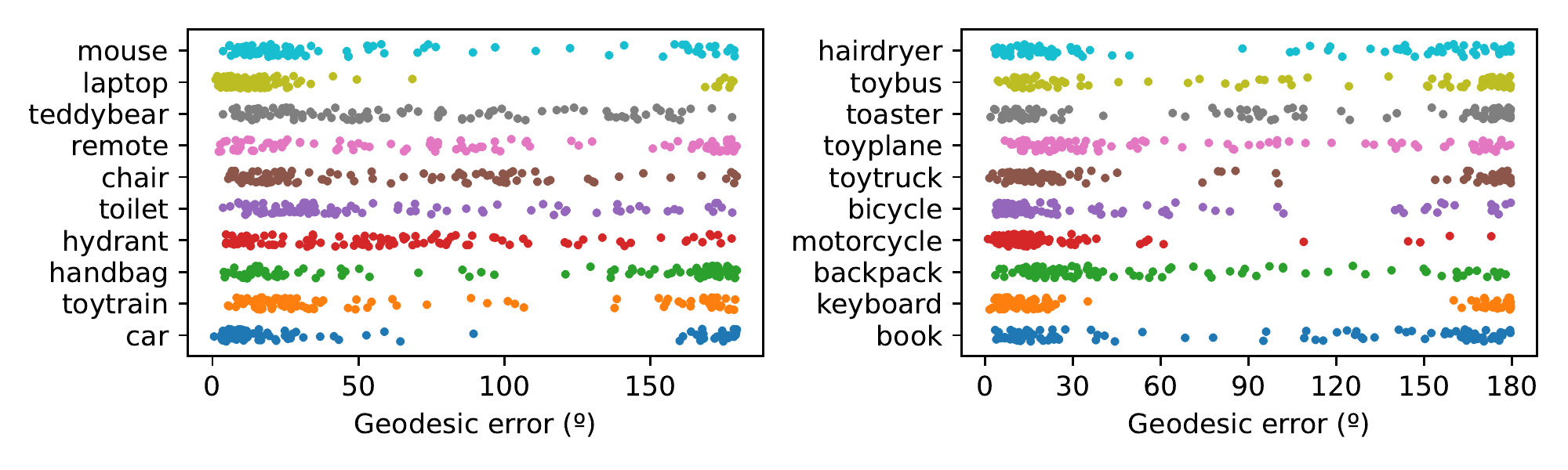}
  \caption{The results from 100 pose estimation problems for each of the 20 categories considered (for the 10 view setting considered in \cref{sec:num_views}). A small amount of vertical displacement is added to the plotted points to make larger clusters salient. For many of the categories, a clear second mode is observed towards maximal geodesic error. In CO3D, where objects tend to vary mostly by an azimuthal rotation (about the gravitational axis), this often reflects a failure mode of predicting `back-to-front' pose for objects that almost exhibit a rotational symmetry between the 0º and 180º azimuthal views (e.g. \textit{bicycles, cars, keyboards, handbags}).}
  \label{fig:err_distrib}
\end{figure}

Many of the 20 categories included in our evaluation exhibit \textit{near} rotational symmetry between a 0º and 180º azimuthal view (about the gravitational axis). For instance, most instances in the \textit{handbag} category have almost complete rotational and mirror symmetry in this sense, with labelling using cues from the handle position to disambiguate pose (see \cref{sec:cat_choice}). To inspect the extent to which categories such as this affect our results, which as just discussed use metrics that enforce a single correct pose label, we plot geodesic errors in 3D orientation prediction from our method in \cref{fig:err_distrib}. Inspect these results, it can be seen that categories that intuitively have a near-symmetry at a 180º offset do tend indeed exhibit a strong bi-modal prediction error that reflects this. For the \textit{chair} and \textit{toaster} categories, where some instances further have 90º rotational symmetry, a third mode of error can be seen that reflects this, also.

\section{Baselines}

\subsection{Iterative closest point}
Iterative Closest Point (ICP) methods seek to minimise the distance between a reference and target point cloud, in the absence of known correspondences, by computing the optimal rigid body transform between these clouds, in an iterative manner. We use the implementation of ICP in the Pytorch3D library \cite{Pytorch3d}, and include a uniform scaling parameter, to match our method's setting. The time complexity of ICP in the number of points $n$ is $O(n ^{2})$, and in order to keep the run-time tractable, we sub-sample each object's point cloud at random to 5000 points prior to running ICP. For the reference object, we construct a point cloud by back-projecting the single reference image using its depth map. For the target object, if multiple views are available, we fuse all of these using the known extrinsics for a more complete point cloud. We use the labelled foreground masks provided in CO3D to produce a masked point cloud - we do not use this in our method except to take a region of interest crop. 

As discussed in \cref{sec:baselines}, we try running ICP both without any pose initialisation (\textbf{ICP}), and - in the multiple target view settings - with initialisation given by the predicted `best frame' from our method. When running without initialisation, we first transform the point clouds to put them in a coordinate frame induced by assuming that the viewing camera (in the reference frame, or in the first frame of the target sequence) is in the position of the camera in first frame of the sequence. That is, for the $i$\textsuperscript{th} reference frame $\text{ref}_{i}$, we transform the reference point cloud by $\mathbf{T}^{\text{cam}}_{\text{ref}_{0}} \circ (\mathbf{T}^{\text{cam}}_{\text{ref}_{i}})^{-1}$, where $\mathbf{T}^{\text{cam}}$ denotes a world-to-view camera transform, and $\text{ref}_{0}$ is the first frame in the reference sequence. This is to reduce a bias in CO3D towards point clouds that are very nearly already aligned in their standard coordinate frames - the camera extrinsic orientation is always the same in first frame of each sequence, and the point cloud coordinate frame is defined with respect to this. For most categories, the crowd-sourced videos start from a very similar viewpoint, which leads to nearly aligned point clouds. When initialising from a best-frame estimate with index $j^{*}$, we use this frame's extrinsics to transform the reference point cloud i.e. $\mathbf{T}^{\text{cam}}_{\text{ref}_{0}} \circ (\mathbf{T}^{\text{cam}}_{\text{ref}_{j*}})^{-1}$ to bring it in line with this view. 

\subsection{PoseContrast}
PoseContrast \cite{Xiao2021} is an RGB-based method designed for zero-shot category level 3D pose estimation. In contrast to our work, it only estimates SO(3) pose, with no translation estimate. It makes use of a pre-trained ResNet50 backbone, and trains on pose-labelled category-level datasets (Pascal3D \cite{Xiang2014} and Objectnet3D \cite{Xiang2016}) with a contrastive loss based on the geodesic difference in pose between samples. Intuitively, it seeks to learn an embedding space in which objects of similar pose are closer together, in the hope that this will generalise to previously unseen categories. The authors note that zero-shot successes are still only probable in cases in which the unseen category has both similar appearance, geometry and canonical reference frame to a category in the training set. As canonical reference frames can be arbitrarily chosen, this makes the success or otherwise of this method entirely dependent on a dataset's choice for category reference frames. In the present work, we formulate zero-shot pose as agnostic of canonical frame, by providing the reference frame implicitly through use of a single reference image. To directly compare to PoseContrast, we bring PoseContrast to the relative setting too. First, PoseContrast estimates a 3D pose for both reference and target frames individually. We then compute the relative SO(3) transform between these two estimates to form the final prediction. We then compare this to the ground-truth given by our labelling process as in all other methods. 
\end{document}